\documentclass[review]{elsarticle}

\usepackage{hyperref}

\usepackage{graphicx}
\usepackage{setspace}
\usepackage{multicol}
\usepackage{multirow}
\usepackage{booktabs}
\usepackage{rotating}
\usepackage{color}
\usepackage[dvipsnames]{xcolor} 
\usepackage{tabularx}
\usepackage{caption}
\usepackage{here}
\usepackage{wrapfig}
\usepackage{algpseudocode}
\usepackage{algorithmicx}
\usepackage{algorithm}
\usepackage{amsmath, amssymb}
\usepackage{pbox}
\usepackage{capt-of}
\usepackage{enumitem}

\usepackage[bottom]{footmisc}
\usepackage{tikz}
\tikzset{
  treenode/.style = {shape=rectangle, rounded corners,
                     draw, align=center,
                     top color=white, font=\small},
  root/.style     = {treenode, font=\small},
  env/.style      = {treenode, font=\small},
  dummy/.style    = {circle,draw}
}

\journal{}
\date{}

\usepackage{lipsum}
\makeatletter
\def\ps@pprintTitle{%
 \let\@oddhead\@empty
 \let\@evenhead\@empty
 \def\@oddfoot{}%
 \let\@evenfoot\@oddfoot}
\makeatother

\begin{document}

\begin{frontmatter}

\title{A One-Class Classification Decision Tree \\ Based on Kernel Density Estimation}

\author{Sarah Itani\fnref{label1,label2}\corref{cor}}
\ead{sarah.itani@umons.ac.be}

\author{Fabian Lecron\fnref{label3}}
\ead{fabian.lecron@umons.ac.be}

\author{Philippe Fortemps\fnref{label3}}
\ead{philippe.fortemps@umons.ac.be}

\cortext[cor]{\raggedright Corresponding author. University of Mons, Department of Mathematics and Operations Research, Rue de Houdain, 9, 7000 Mons, Belgium.}
\address[label1]{Fund for Scientific Research - FNRS (F.R.S.-FNRS), Brussels, Belgium} 
\address[label2]{Faculty of Engineering, University of Mons, Department of Mathematics and Operations Research, Mons, Belgium} 
\address[label3]{Faculty of Engineering, University of Mons, Department of Engineering Innovation Management, Mons, Belgium} 

\begin{abstract}
One-class Classification (OCC) is an area of machine learning which addresses prediction based on unbalanced datasets. Basically, OCC algorithms achieve training by means of a single class sample, with potentially some additional counter-examples. The current OCC models give satisfaction in terms of performance, but there is an increasing need for the development of interpretable models. In the present work, we propose a one-class model which addresses concerns of both performance and interpretability. Our hybrid OCC method relies on density estimation as part of a tree-based learning algorithm, called One-Class decision Tree (OC-Tree). Within a greedy and recursive approach, our proposal rests on kernel density estimation to split a data subset on the basis of one or several intervals of interest. Thus, the OC-Tree encloses data within hyper-rectangles of interest which can be described by a set of rules. Against state-of-the-art methods such as Cluster Support Vector Data Description (ClusterSVDD), One-Class Support Vector Machine (OCSVM) and isolation Forest (iForest), the OC-Tree performs favorably on a range of benchmark datasets. Furthermore, we propose a real medical application for which the OC-Tree has demonstrated its effectiveness, through the ability to tackle interpretable diagnosis aid based on unbalanced datasets.
\end{abstract}


\end{frontmatter}

\section{Introduction}
As precious assets of knowledge extraction, data are massively collected in the fields of industry and research, day by day. Though valuable, the proliferation of data requires attention upon processing. In particular, unbalanced datasets may be hardly addressed through the classical scheme of multi-class prediction. The practice of One-Class Classification (OCC) has been developed within this consideration~\cite{Moya1993, Khan2009}. 

OCC is of major concern in several domains where it may be expensive and/or technically difficult to collect data on a range of behaviors or phenomenons~\cite{Chandola2009}. For example, it may be quite affordable to gather data on the representatives of a given pathology in medicine, or positive operating scenarios of machines in the industry. The related complementary occurrences are, by contrast, scarce and/or expensive to raise~\cite{Khan2009}. As a matter of fact, one-class classifiers are trained on a single class sample, in the possible presence of a few counter-examples. The resulting models allow to predict \textit{target} (or \textit{positive}) patterns and to reject \textit{outlier} (or \textit{negative}) ones. Basically, OCC is pursued for outlier (or anomaly) detection. 

One-Class Support Vector Machine (OCSVM) and Support Vector Data Description (SVDD) are among the most common OCC methods~\cite{Scholkopf2001, Tax2001}. OCSVM aims at finding the hyper-plane that separates the target instances from the origin with the wider margin, while SVDD aims at enclosing these instances within a single hyper-sphere of minimal volume. Far from being contested, the effectiveness of these methods has notably been improved with the development of variants that better fit some data structures~\cite{Yu2005, Wu2009, Xiao2009, Le2010b, Nguyen2012, Yang2016}. Indeed, the instances of a single class may be enclosed within several groupings in the form of \textit{sub-concepts} that it would be interesting to raise separately~\cite{Barnabe2015}. ClusterSVDD~\cite{Gornitz2017} achieves such a purpose: this recent method may be seen as a $K$-means algorithm~\cite{Hartigan1975} ruled by the results of distinct SVDD problems.

Admittedly, the current methods of OCC give satisfaction, but that is without counting on the advent of explainable artificial intelligence which opens new research horizons for machine learning in encouraging the development of interpretable models~\cite{Holzinger2017}. In this regard, some methods have been developed as post-hoc explainers on the predictions of classifiers~\cite{Hara2016, Ribeiro2016}. But a great challenge remains the development of interpretable models by nature, which provide simultaneously high levels of performance. This challenge is the major source of motivation for the present work. 

Originally introduced for supervised classification, decision trees~\cite{Quinlan1986} provide satisfaction in terms of interpretability. The extensions of the algorithm proposed to tackle OCC often rely on the generation of outliers~\cite{Desir2013, Goix2016}. However, a decision tree is basically built under the hypothesis that the different classes cover the whole domain by their representatives. Thus, the one-class variants may associate the target class with a large subspace against the one which the class occupies in reality. In a different perspective, the work of~\cite{Liu2008} revisits the development of decision trees by orienting the training process towards the isolation of outliers rather than of target instances. The intuition behind the method, called Isolation Forest (iForest), is that outliers are scarce and easily detectable compared to target instances~\cite{Liu2008}. The outliers can thus be isolated by means of a low number of divisions. An IF is an ensemble of trees built based on a random choice of attributes and thresholds. For a given instance, if the average path skimmed in the trees is short, the instance is predicted as outlier.


Kernel Density Estimation (KDE)~\cite{Silverman1986} is another approach which can address OCC intuitively, in computing the non-parametric estimation of a sample distribution. Thresholded at a given level of confidence, this estimation is used to reject any instance located beyond the decision boundary thus established. However, KDE loses in performance and readability towards high dimensional samples~\cite{Desir2013}.  

In the present work, we tackle OCC through a hybrid method, called One-Class decision Tree (OC-Tree), which is intended to combine the benefits of the standard decision tree and KDE. 
The contributions of our work are exposed below. 
\begin{sidewaysfigure}
\centering
\scalebox{.8}{
\begin{minipage}[c]{.46\linewidth}
\centering
\includegraphics[scale=0.7]{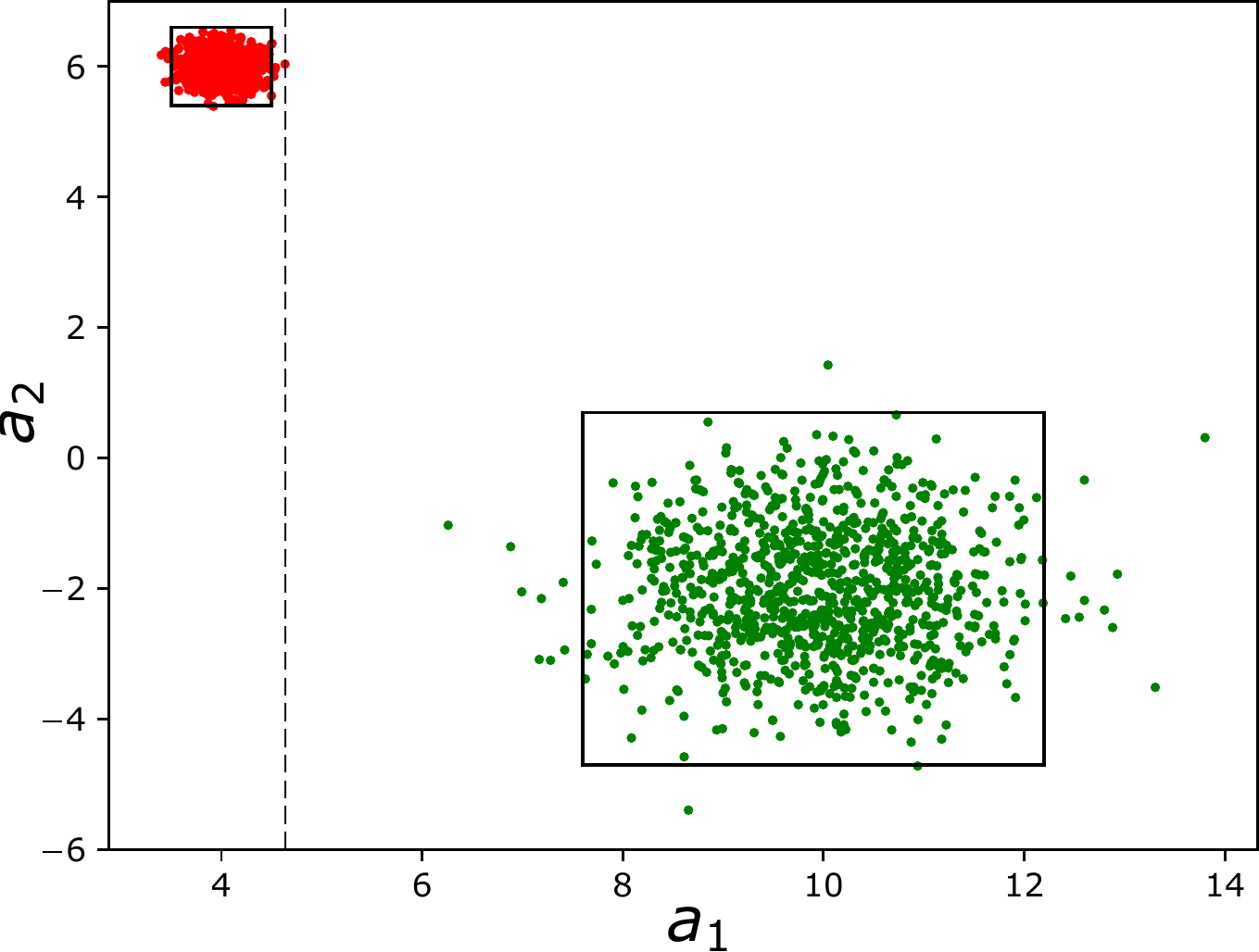}
\end{minipage} \hspace{3cm}
\begin{minipage}[c]{.5\linewidth}
\textbf{Multi-class decision tree} 
\begin{small}
\begin{center}{
\resizebox{0.7\textwidth}{!}{%
\begin{tikzpicture}
  [
    grow                    = down,
     level 1/.style={sibling distance=40mm, level distance = 3em},
     level 2/.style={sibling distance=15mm, level distance = 4em},
     level 3/.style={sibling distance=30mm },
     level 4/.style={sibling distance=30mm},
    edge from parent path={(\tikzparentnode) -- (\tikzchildnode)}],
    every node/.style       = {font=\footnotesize},
    sloped
  ]
  \node [root] {}
        child {node[env]{$C_1$}  
    		edge from parent node [auto=right, pos=.8] {$a_1 \leq 4.6$} }
        child {node[env]{$C_2$}  
    			edge from parent node [auto=left, pos=.8] {$a_1 > 4.6$} };			
\end{tikzpicture}}}\end{center}
\end{small}
\color{white}[0.1cm]\color{black}\\

\textbf{OC-Tree}
\begin{small}
\begin{center}
\begin{tikzpicture}
  [
    grow                    = down,
     level 1/.style={sibling distance=25mm, level distance = 3em},
     level 2/.style={sibling distance=15mm, level distance = 4em},
     level 3/.style={sibling distance=30mm },
     level 4/.style={sibling distance=30mm},
    edge from parent path={(\tikzparentnode) -- (\tikzchildnode)}],
    every node/.style       = {font=\footnotesize},
    sloped
  ]
  \node [root,font=\small] {}
        child {node[env]{}  
                child {node[env]{$C$}  
    				edge from parent node [auto=right, pos=.8] {$a_1 \in [7.6;12.2]$} }
    			child {node[env]{$O$}  
    				edge from parent node [auto=left, pos=.8] {else} }
    		edge from parent node [auto=right, pos=.8] {$a_2 \in [-4.7;0.7]$} }
        child {node[env]{$O$}  
    			edge from parent node [auto=left, pos=.65] {else} }
        child {node[env]{}  
                child {node[env]{$C$}  
    				edge from parent node [auto=right, pos=.8] {$a_1 \in [3.5;4.5]$} }
        		child {node[env]{$O$}  
    				edge from parent node [auto=left, pos=.8] {else} }
    			edge from parent node [auto=left, pos=.8] {$a_2 \in [5.4;6.6]$} };			
\end{tikzpicture}
\end{center}
\end{small}
\end{minipage}}
\caption{Comparison between a multi-class and our one-class decision tree on an artificial dataset}\label{Example}
\end{sidewaysfigure}
\begin{itemize}
\item Compared to previous adaptations of the decision tree to OCC, our proposal rather focuses on the isolation of the target training instances through a density-based hierarchical process of splitting, in which subdivisions are based on closed intervals of interest. This innovative splitting mechanism is supported by KDE. 
\item The combination of decision trees with kernel density estimators was originally proposed by \textit{Smyth et al.}~\cite{Smyth1995}. Such a method was intended to boost the performances of the standard decision tree, in using a kernel density estimator to compute posterior class probabilities, based exclusively on the attributes belonging to a given decision chain. Our proposal differs in some regards. Indeed, the work of~\cite{Smyth1995} tackled multi-class classification, through an approach which is implemented in two phases: (A) decision tree induction and (B) kernel density estimation to compute class probability. Our proposal tackles OCC through a hybrid methodology where density estimation is considered as a part of decision tree induction. 
\item The OC-Tree may be seen as the integration of a multi-dimensional KDE within an intuitive and structured decision scheme. Indeed, the OC-Tree encloses data within hyper-rectangles, based on a subset of training attributes selected for their discriminative power. 
\item The method has shown favorable performances in comparison to reference methods, including ClusterSVDD~\cite{Gornitz2017}, OCSVM~\cite{Scholkopf2001} and iForest~\cite{Liu2008} on benchmark datasets. 
\item We apply our algorithm for the diagnosis of Attention Deficit Hyperactivity Disorder (ADHD), based on the ADHD-200 collection. In this regard, the classification accuracy achieved by the OC-Tree is competitive in comparison to the results reported in the recent literature. We believe that the convenience and the interpretability of the OC-Tree make it a promising model for future clinical practice.
\end{itemize}

To illustrate the objective of our OC-Tree, let us consider a toy example proposed in Fig.~\ref{Example} (left). The latter is processed in two distinct ways : 
\begin{itemize}
\item \textit{with a multi-class decision tree}. In this case, each Gaussian blob is associated to a distinct class ($C_1$, in red and $C2$, in green). The associated space division is represented in dashed lines. 
\item \textit{with an OC-Tree}. In this case, the Gaussian blobs are all the representatives of the same Class (called $C$). The limits of the corresponding hyper-rectangles are represented in continuous lines. The complementary space is the one of Outliers (called $O$). 
\end{itemize} 
As shown, multi- and one-class learning processes lead to different predictive models (Fig.~\ref{Example}, right). Indeed, in the context of a multi-class problem, the class representatives are supposed to share the whole domain in which the attributes take their values. Hence, a decision tree learned with an algorithm like C4.5~\cite{Quinlan1986, Quinlan1993} proposes a decomposition of the whole space in hyper-rectangles by means of one single attribute. On the opposite, aiming at solving a one-class classification problem, we propose a learning process looking for target hyper-rectangles that do not necessarily cover the whole domain in which the attributes take their values, since there may exist outliers to discard. 

The remainder of the paper is organized as follows. Sec.~\ref{RelatedWork} describes our algorithm which was assessed in comparison to reference methods according to the experimental protocol presented in Sec.~\ref{ExpPro}. We expose the related results in Sec.~\ref{Results}. Then, in Sec.~\ref{CaseStudy}, we present a medical case study whose challenging aspects can be appropriately addressed by the OC-Tree. Finally, we discuss and summarize our findings in Sec.~\ref{Discussion}, before concluding the paper in Sec.~\ref{Conclusion}.

\section{Our proposal}\label{RelatedWork}

In a \textit{divide and conquer} spirit, the implementation of our one-class tree rests on successive density estimations to raise target areas as hyper-rectangles of interest. We assess the relevance of a subdivision against an information gain criterion adapted to OCC issues proposed by~\cite{Goix2016}.

Let us consider $\chi \subset \mathbb{R}^{d}$ a hyper-rectangle of dimensions $d$ including target training instances. Let us note $A = \lbrace a_1, a_2,\ldots, a_d \rbrace$ the set of attributes and $X = \lbrace x_1, x_2,\ldots, x_{n}\rbrace$ the set of instances. The goal of our proposition is the division of the initial hyper-rectangle $\chi$ in (non necessarily adjacent) sub-spaces $\chi_{t_i}$, represented by tree nodes $t_i$, in absence of counter-examples.  

Let us denote as $A_t$ the set of eligible attributes for division at a given node $t$. Thus, $A_t \subseteq A$. We note $A_t = \lbrace a'_{1}, a'_{2},\ldots a'_{l_t}\rbrace$, $l_t$ being the number of eligible attributes at node $t$, with $l_t \leq d$ accordingly.  At each node $t$, the algorithm searches the attribute $a'_{j} \in A_t$ which best cuts the initial sub-space  $\chi_{t}$ into one or several sub-space(s) $\chi_{t_i}$ such that:
\begin{equation}
\chi_{t_i} = \lbrace x \in \chi_{t} : L_{t_i}\leq x^{a'_{j}}\leq R_{t_i}  \rbrace
\label{childNodes}
\end{equation}
$x^{a'_{j}}$ is the value of instance $x$ for attribute $a'_{j}$; $L_{t_i}$ and $R_{t_i}$ are respectively the left and right bounds of the closed sub-intervals raised to split the current node $t$ in target nodes $t_i$, based on attribute $a'_{j}$. 

For each attribute $a'_{j} \in A_{t}$, the algorithm achieves the following steps, at a given node $t$.
\begin{enumerate}
\item[1.] Check if the attribute is still eligible and compute the related Kernel Density Estimation (KDE), i.e., an estimation of the probability density function $\hat{f}_j(x)$ based on the available training instances (see Sec.~\ref{KDE}). 
\item[2.] Divide the space $\chi_{t}$, based on the modes of $\hat{f}_j(x)$ (see Sec.~\ref{Division}). 
\item[3.] The quality of the division is assessed by the computation of the \textit{impurity} of the resulting nodes deriving from division (see Sec.~\ref{Assess}). 
\end{enumerate} 
At each iteration, the attribute that achieves the best purity score is selected to split the current node $t$ in child nodes. If necessary, some branches are pre-pruned in order to preserve the interpretability of the tree (see Sec.~\ref{Prepruning}). The algorithm is run recursively; termination occurs under some stopping conditions (see Sec.~\ref{Stop&Eligibility}). 

In the rest of this paper, what we refer to as the \textit{training accuracy} corresponds to the rate of training instances included in target nodes. It follows that, in this context of OCC, the training classification error corresponds to the rate of training instances predicted as outliers by the predictive model. 
\subsection{Density estimation}\label{KDE}
In order to identify concentrations of target instances, we have to estimate their distribution over the space, which can be provided by a Kernel Density Estimation (KDE). In particular, our proposal is based on the popular Gaussian kernel~\cite{Silverman1986}:
\begin{equation*}
\hat{f}_j(z) = \frac{1}{n_t h_t}\sum_{i = 1}^{n_t}K\left(\frac{z-x_i}{h_t}\right)  \:\:\:\:\text{with} \:\:\:\: K(y) = \frac{1}{\sqrt{2\pi}}\exp{\frac{-y^2}{2}}
\end{equation*}
where $\hat{f}_j$ is the KDE related to attribute $a'_{j}$, $X_t = \lbrace x_1, x_2,\ldots, x_{n_t}\rbrace$ is the set of $n_t$ instances available at node $t$, $K$ the kernel function and $h_t$, a parameter called \textit{bandwidth}. 

The parameter $h_t$ influences the pace of the resulting function $\hat{f}_j(x)$~\cite{Silverman1986}. As $h_t$ tends towards zero, $\hat{f}_j(x)$ appears over-shaped while high values of $h_t$ induce a less detailed density estimation. Adaptive methods, such as a least-squares cross-validation, may help setting the bandwidth value~\cite{Jones1996, Li2007}. However, such iterative techniques are computationally expensive; their use may be hardly considered in this context of recursive divisions. Hence, we compute $h_t$ as~\cite{Silverman1986}:
\begin{align}
	h_t= 
   \begin{cases} 
   0.9\cdot  \min(\hat{\sigma}, IQR/1.34)\cdot n_t^{-1/5} &\text{if} \:\ IQR \neq 0\\
   0.9\cdot \hat{\sigma}\cdot n_t^{-1/5} &\text{otherwise}\\
   \end{cases}
\label{bandwidth2}
\end{align}
where $\hat{\sigma}$ is the standard deviation of the sample $X_t$ and $IQR$, the associated inter-quartile range. 
The first relation corresponds to the Silverman's \textit{rule of thumb}~\cite{Silverman1986}. We consider the second relation to address samples with $IQR=0$, which may reveal very concentrated data, with the potential presence of some singularities that should be eliminated.


\subsection{Division}\label{Division}
\begin{figure}[t]
\centering
\includegraphics[height=0.8\textwidth]{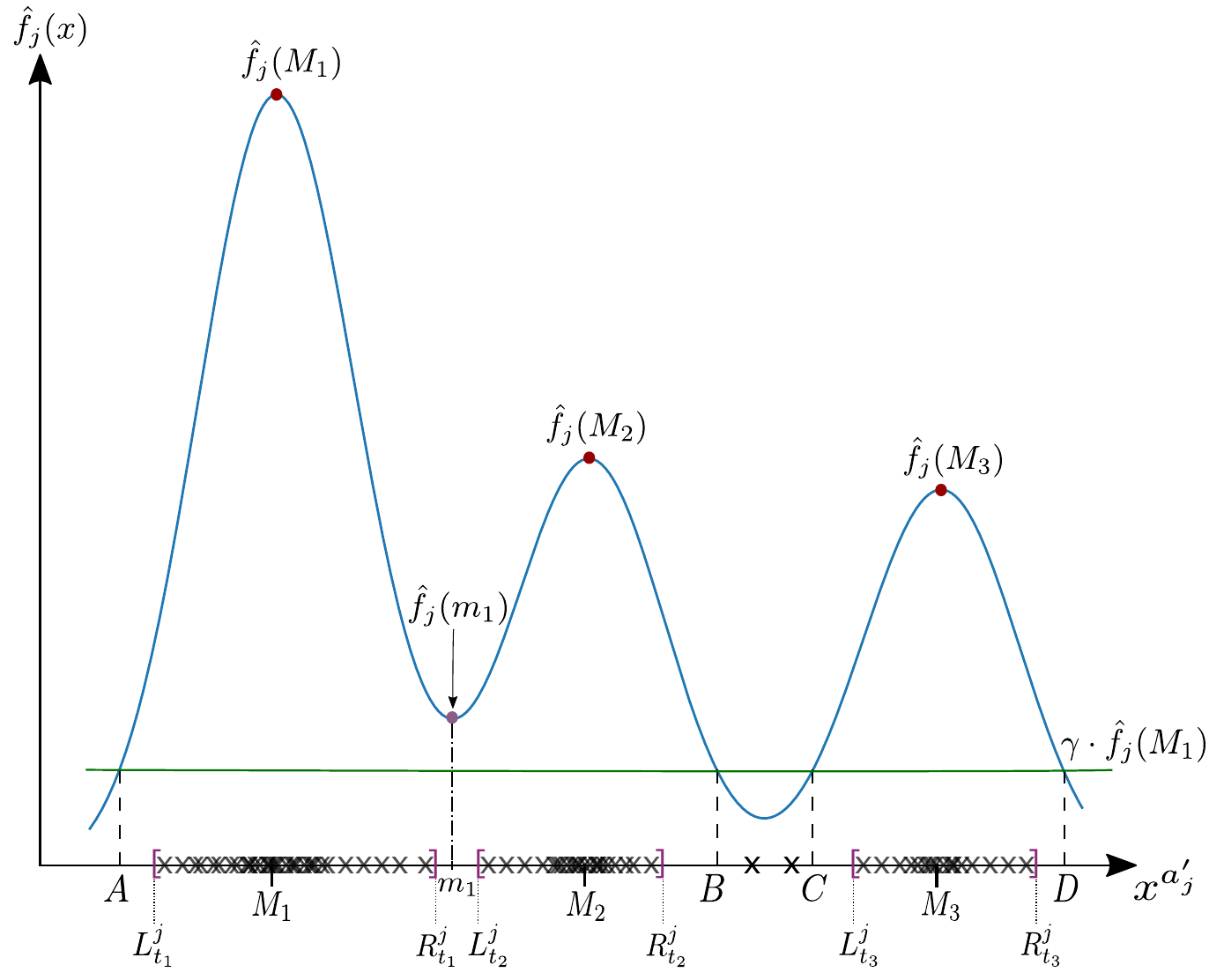}
\caption{Division mechanism}\label{ReviseSplit}
\end{figure}
At node $t$, division is executed based on $\hat{f}_j(x)$, in four steps. 
\begin{itemize}
\item [(a)] \textbf{Clipping KDE ($\gamma$)}\\ $\hat{f}_j(x)$ is thresholded at the level $\gamma\cdot \max_{x\in \chi_t} \hat{f}_j(x)$. \\
This allows to raise a set of target sub-intervals ${Y_j}^{t}$. 
\item [(b)] \textbf{Revision ($\alpha$)}\\
If $\hat{f}_j(x)$ is $k$-modal ($k\neq1$) and $1\leq\vert {Y_j}^{t}\vert < k$, revision occurs since some modes were not identified. Each sub-interval of ${Y_j}^{t}$ is thus analyzed: if its image by $\hat{f}_j(x)$ includes at least a significant local minimum, the interval is split in two sub-intervals around this (these) local minimum (minima). The significance of a local minimum is assessed through a parameter $\alpha$ (see below). 
\item [(c)] \textbf{Assessment ($\beta$)}\\
The sub-intervals of ${Y_j}^{t}$ covering a number of training instances inferior to a quantity $\beta.\vert T\vert$ are dropped. This ensures keeping the most significant target nodes.
\item [(d)] \textbf{Shrinking}\\
The detected sub-intervals are shrunk in closed intervals in a way to fit the domain strictly covered by the related target training instances, as defined by Eq.~\ref{childNodes}.
\end{itemize}
Actually, ${Y_j}^{t}$ may be updated at the end of steps (b), (c), (d). 

If we consider the KDE presented by Fig.~\ref{ReviseSplit}, (a) results in ${Y_j}^{t} = \lbrace [A, B]; [C, D] \rbrace$. As the density estimation is 3-modal in this case, a revision of the interval partitioning (b) is launched. It appears there is no need to split the sub-interval $[C, D]$ since the piecewise $\hat{f}_j([C,D])$ includes a single maximum. By contrast, a local minimum is detected in $m_1$, in the piecewise $\hat{f}_j([A,B])$. The sub-interval [A,B] is thus split into three parts around the local minimum. 
\noindent
Concretely, such a split occurs if the local minimum is significant, i.e., sufficiently deep in comparison with both nearby local maxima. In mathematical terms:
\begin{equation*}
\hat{f}_j(m_1) \leq \alpha \cdot \min(\hat{f}_j(M_1), \hat{f}_j(M_2)).
\end{equation*} 
Thus  ${Y_j}^{t} = \lbrace [A, m_1[; [m_1,C[; [C, D] \rbrace$. Steps (c) and (d) are then launched. The sub-intervals are shrunk around the target training instances (represented by crosses in Fig.\ref{ReviseSplit}), which results in:
\begin{equation*}
{Y_j}^{t} =  \lbrace [L_{t_1}^j, R_{t_1}^j];  [L_{t_2}^j, R_{t_2}^j] ; [L_{t_3}^j, R_{t_3}^j] \rbrace.
\end{equation*}
The complement $\overline{{Y_j}^{t}}$ represents the set of outlier sub-spaces: it may be represented by a single branch entitled "else".  

Except for prior knowledge that would help choosing its value more specifically, there should be no reason to set a high reject threshold \label{parameters}$\beta$ (e.g., $> 2\%$) since the training set is supposed to include a majority of target instances; this would be penalizing with the exclusion of real target nodes as a consequence. An appropriate value for parameter $\alpha$ may be selected by cross-validation; actually, a non-zero value for $\alpha$ (e.g., 0.5) will lead to revision, which appears to be interesting if we want to detect precisely target groupings.  Basically, the value of the clipping threshold $\gamma$ should be low (e.g., 0.05), because it aims at rejecting outliers. 

\subsection{Impurity decrease computation}\label{Assess}

At this stage of the algorithm, we have to assess the quality of a division in a particular context, i.e., the absence of representatives for at least a second class. One way to achieve this task is to resort to the physical generation of $n'_t$ outliers in each node~\cite{Hempstalk2008, Desir2013}; as a result of the division, each child node would include a number of $n'_{t_i}$ instances which would have to be estimated. 

The virtual generation of outliers is worth considering as well. In this regard, the work of~\cite{Goix2016} assumes each parent node includes uniformly distributed outliers in equal number to that of the target instances, i.e., $n_t = n'_t$. Thus, the number of outliers in each child node may be easily deduced: 
\begin{equation}
n'_{t_i} = n'_{t}\frac{\mu(\chi_{t_i})}{\mu(\chi_{t})}
\label{equation}
\end{equation}
where $\mu$ denotes the measure of the hyper-rectangle to which it relates.

Assuming $n_t=n'_t$ may appear counter-intuitive. Indeed, we would naturally be inclined to assume, once and for all, $n = n'$ in the initial root node and to deduce the number of outliers in each child node according to Eq.~\ref{equation}. But throughout the iterations, this would lead to increase the scarcity of the outliers, and thus to their unfair representation in each node. The latter situation corresponds to the well-known effect of the \textit{curse of dimensionality}~\cite{Bellman1957}. This is why, to address this issue, the number of outliers in each node $t$ is considered as corresponding to the number $n_t$ of target instances prior to any division~\cite{Goix2016}. 

Based on this predictive calculation, the work of \cite{Goix2016} gives a proxy for the Gini impurity decrease for OCC. We adapt this result to our proposal where a division may result in more than two child nodes $t_i$, based on sub-intervals of interest:
\begin{align*}
I_G^{Proxy} (t_i, 1\leq i \leq r_t) = \sum_{i=1}^{m_t} \frac{n_{t_i}n'_{t_i}}{n_{t_i}+n'_{t_i}} \\
\text{with }\:\:\: n'_{t_i} = n'_{t}\frac{R^{a'_j}_{t_i} -  L^{a'_j}_{t_i}}{R^{a'_j}_{t} -  L^{a'_j}_{t}} 
\end{align*}
where $r_t$ is the total number of target and outlier sub-intervals, included in ${Y_j}^{t} \cup \overline{{Y_j}^{t}}$. 

\begin{figure}
\centering
\includegraphics[width=0.9\textwidth]{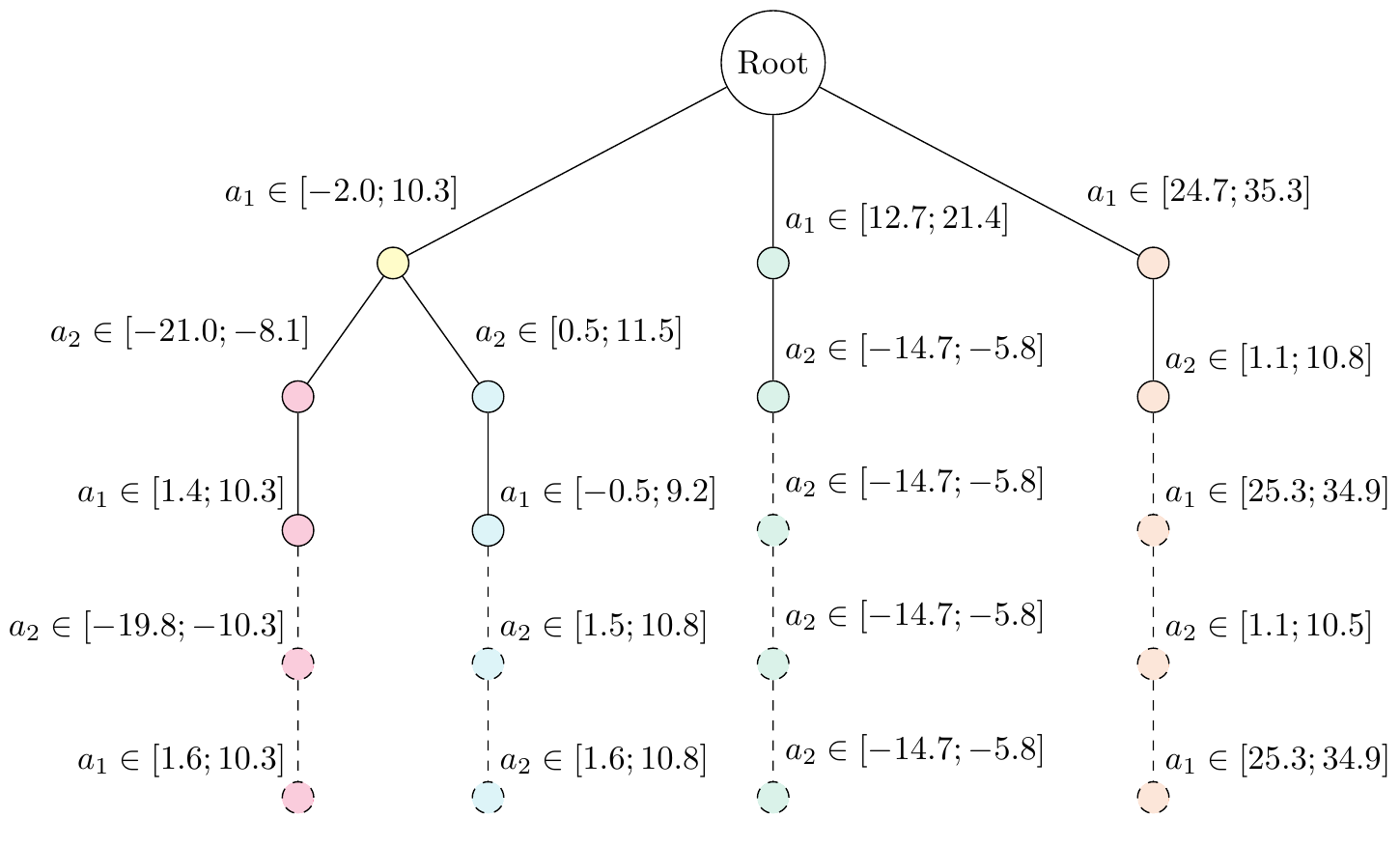}
\caption{Pre-pruning mechanism}\label{prepruning}
\end{figure}
\subsection{Pre-pruning mechanism}\label{Prepruning}

A branch of an OC-Tree is prepruned if there are no more eligible attributes for division. An attribute is not eligible if:
\begin{itemize}
\item for this attribute, all the instances have the same value;
\item the attribute was already used previously to cut the same target node which was not split in several target nodes in the meantime; 
\item the computed bandwidth $h_t$ is strictly inferior to the minimum of the difference between two (different) successive values in the set of available instances, i.e., data granularity. 
\end{itemize}
At a given node $t$, a division based on a non-eligible attribute makes no more sense. 
Fig.~\ref{prepruning} shows a tree trained on two attributes. The nodes in dotted lines are developed in absence of a pre-pruning mechanism; the latter allows to get a shorter and readable decision tree. Note that the branches related to outliers were omitted for the sake of clarity. 

The user has basically the choice to keep either the tree as (1) a full predictive model which describes the development that brought to the space division, or (2) the description of the final target hyper-rectangles as a set of sub-intervals of interest regarding the attributes that were used for division. 

\subsection{Stopping conditions}\label{Stop&Eligibility}
Let us denote the training accuracy as $A_{tr}$: it corresponds to the ratio of training instances included in the target nodes. The algorithm stops under some global and local conditions. 

\begin{itemize}
\item \textbf{Globally}, the algorithm is stopped: 
\begin{itemize}
\item  if $A_{tr}$ remains stable after an iteration in which no additional target node was raised. In this case, the training process reaches a stage where the target sub-spaces are simply more precisely delimited on the basis of additional attributes, with no further multiplication.  
\item if $A_{tr} < 1-\nu$, where $\nu$ is a parameter corresponding to the fraction of training instances which we tolerate to reject and to consider as outliers. 
\end{itemize}
\item Divisions may be stopped \textbf{locally} if there are compelling reasons to convert a node in a leaf, i.e., when pre-pruning is necessary (see Sec.~\ref{Prepruning}). 
\end{itemize}

\section{Experimental protocol}\label{ExpPro}
\begin{figure*}
\centering
\includegraphics[scale=0.4]{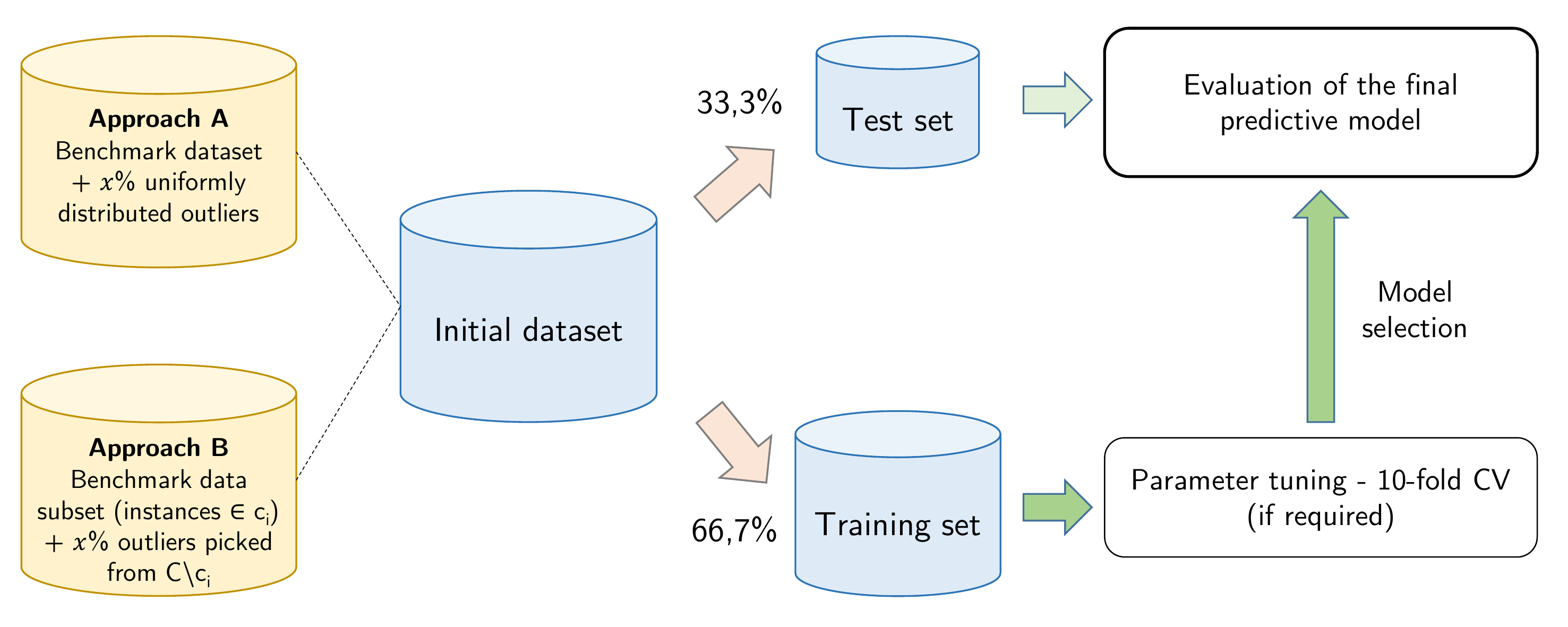}
\caption{Experimental pipeline}\label{Pipeline}
\end{figure*}

Figure~\ref{Pipeline} summarizes our experimental protocol which is explained in detail in the following sections.  
\subsection{Reference methods}
We compared the OC-Tree with three reference methods, namely the ClusterSVDD~\cite{Gornitz2017}, One-Class Support Vector Machine (OCSVM)~\cite{Scholkopf2001} and Isolation Forest (iForest)~\cite{Liu2008}. 
 
The comparison of the OC-Tree with ClusterSVDD is highly relevant since both methods pursue similar objectives, i.e., enclosing data within one or several hyper-rectangle(s) and hyper-sphere(s) respectively. ClusterSVDD requires that two parameters should be optimized on a dataset: $\nu$ and $k$ which constitute respectively, the upper bound on the fraction of instances lying outside the decision boundary and the supposed number of clusters. Table~\ref{Comp} exposes a theoretical comparison of the OC-Tree with ClusterSVDD. 

OCSVM is a standard OCC method to which a comparison is thus worth considering. We considered a gaussian kernel for this method, and we optimized $\nu$ which pursues the same objective as in ClusterSVDD and OC-Tree. Thus, to ensure a fair comparison, we adjusted this parameter in the same way that we did for ClusterSVDD. Finally, a method like iForest provides a relevant benchmark since it is of the same nature than OC-Tree, i.e., a tree-based method, but built in a very different way. Indeed, this ensemble technique aims at the development of decision trees based on a random choice of attributes and thresholds. If the average path length skimmed in the trees is low (resp. high), an instance is predicted as outlier (resp. target). We used the standard parameter settings for this method, since it was shown that the performances are ensured to be quite optimal with such settings~\cite{Liu2008}. 

\begin{table}
\centering
\resizebox{\textwidth}{!}{
\begin{tabular}{l|l}
\parbox{6.5cm}{\centering \textbf{ClusterSVDD}} & \multicolumn{1}{c}{\textbf{OC-Tree}} \\ \toprule
\parbox{6.5cm}{\begin{itemize}
\item Detects target hyper-sphere(s).
\item Requires to set the number of hyper-sphere(s) as a parameter.
\item Relies on two parameters: $k$, $\nu_{SVDD}$.
\item Results in a classification model whose predictions are based on the whole set of training attributes.
\end{itemize}} & 
\parbox{6.5cm}{\begin{itemize}
\item Detects target hyper-rectangle(s).
\item Does not require indications about the number of hyper-rectangle(s) to detect. 
\item Relies on four parameters: $\gamma, \beta, \alpha, \nu$.
\item Results in a classification model whose predictions are based on a subset of training attributes.
\end{itemize} }\\
\bottomrule
\end{tabular}}
\caption{Comparison of ClusterSVDD \& OC-Tree}\label{Comp}
\end{table}

\subsection{Benchmark datasets} \label{secData}

In absence of benchmark data for OCC, it is standard practice to convert multi-class problems into one-class ones for evaluation purposes. We thus considered a set of benchmark datasets (see Table~\ref{Data}), where each instance belongs to a class $c_i$ among a set of $C$. The relevancy of OC-Tree and of the reference methods on these datasets was assessed in two distinct ways. 
\begin{itemize}
\item[A.] All the instances, whatever their class, were considered as the representatives of a same class. We injected in this dataset a certain percentage of additional outliers following a uniform distribution~\cite{Gornitz2017}. (\textbf{Approach A})
\item[B.] We adopted the \textit{one vs rest} \cite{Desir2013} strategy which consists of considering a class $c_i \in C$ as a target one and the others as outliers~\cite{Wang2006, Hempstalk2008, Desir2013, Nguyen2015, Fragoso2016, Wang20162}. In this case, the outliers injected in a given data subset were randomly picked among the representatives of the 	outlier classes, i.e., $C\setminus c_i$. (\textbf{Approach B}) 
\end{itemize}
\begin{table}
\centering
\begin{tabular}{l|c|c|c}
& \# Classes& \# Features& \# Instances \\ \toprule 
Australian & 2 & 14& 690 \\ [0.05cm]
Diabetes & 2 & 8 & 268 \\ [0.05cm]
Ionosphere & 2 & 34 & 351 \\ [0.05cm]
Iris & 3 & 4 & 150 \\  [0.05cm]
Satimage & 6 & 36 & 4435 \\ [0.05cm]
Segment & 7  &19&2310 \\
\bottomrule
\end{tabular}
\caption{Benchmark datasets~\cite{UCIRepository, Chang2011}}\label{Data}
\end{table}
Whether through approach A or B, the resulting dataset was split in a way that two thirds constituted a training set, while the remaining was kept as a test set.  

\subsection{Evaluation metrics}\label{AssessmentMet}
As one-class classification deals with unbalanced datasets, we may hardly consider true positives (or true targets) and true negatives (or true outliers) as equally significant. On this regard, the couple \textit{precision-recall} provides appropriate evaluation metrics~\cite{Scikit-learn}. 

Let us denote as $TT$ (resp. $TO$), the number of True Targets (resp. True Outliers), i.e., the number of instances correctly detected as targets (resp. outliers); $FT$ (resp. $FO$) are the number of False Targets (resp. False Outliers) \cite{Nguyen2015}. Precision (P) and Recall (R) are defined as follows. 

\begin{itemize}
\item Precision expresses the ratio of instances that were correctly predicted as target ones to those which were predicted as such.  
\begin{equation}
P = \frac{TT}{TT+FT}
\label{P}
\end{equation}
\item Recall expresses the ratio of instances that were correctly predicted as target ones to those which are truly target instances.  
\begin{equation}
R = \frac{TT}{TT+FO}
\label{R}
\end{equation}
\end{itemize}
Precision and recall can be embedded in a single performance indicator, namely the F1-score~\cite{Scikit-learn}.
\begin{equation}\label{F1score}
F1 = 2 \frac{P\cdot R}{P + R}
\end{equation} 

\subsection{Model selection}\label{ModelSelection}
The OC-Tree and some reference methods rely on a certain number of parameters that have to be adjusted appropriately. This parameter tuning was achieved through a 10-fold Cross-Validation (10-fold CV) procedure, based on the values presented in Table~\ref{ParamSettings}. Note that we conducted a grid search in the case where we had to optimize two parameters. 

\begin{table}[t]
\centering
\begin{tabular}{lp{6cm}}
\textsc{Method} & \textsc{Settings} \\ \toprule
\multirow{4}{*}{OC-Tree} & 
\begin{itemize}
\item[$\bullet$] $\gamma = 0.05$
\item[$\bullet$] $\alpha = \lbrace 0.5, 0.6, 0.7, 0.8 \rbrace$
\item[$\bullet$] $ \beta = 2\%$ (min. 5 inst./node) 
\item[$\bullet$] $\nu = \lbrace 0.05, 0.1, 0.15, 0.2 \rbrace$
\end{itemize} 
\\ 
\multirow{2}{*}{ClusterSVDD}  & \begin{itemize}
\item[$\bullet$] $k$ (see Table~\ref{Tuning})
\item[$\bullet$] $\nu = \lbrace 0.05, 0.1, 0.15, 0.2 \rbrace$
\end{itemize} \\
OCSVM & $\nu = \lbrace 0.05, 0.1, 0.15, 0.2 \rbrace$  \\ 
iForest & Not required \\ \bottomrule
\end{tabular}
\caption{Parameter settings}~\label{ParamSettings}
\end{table}

The range of values for parameter $k$, i.e., the number of clusters in ClusterSVDD, has been differentiated depending on the considered dataset and the approach under which the datasets were addressed, as defined in Sec.~\ref{secData}. Some values are suggested in~\cite{Gornitz2017}. More particularly in regards to approach B, it appeared to us reasonable to set a range of $[1,5]$ as possible values for parameter $k$, regardless of the considered dataset. Indeed, in this case, each class of the multi-class problem is considered for OCC. Thus, intuitively, one would expect that data are concentrated within a small number of target groupings but in the same time, the presence of a single class may reveal a structure of data different from the one observed in the case of a multi-class problem. That is why $k$ may present higher values than those considered with approach A for some datasets. 

Thus, except for iForest, each algorithm was tuned through a CV procedure, in quest of the model which presents the best performance at the sense of the F1-score (see Eq.~\ref{F1score}). The model selection was naturally achieved on the training set extracted from each dataset.  The selected models were finally assessed against a test set. 
\begin{table}
\centering
\begin{tabular}{lcc}
& \multicolumn{1}{c}{Approach A} & Approach B \\  \toprule 
Australian & $\lbrace 1, 2 \rbrace $ &   \multirow{6}{*}{$\lbrace 1, 2, 3, 4, 5 \rbrace $}  \\ [0.05cm]
Diabetes &  $\lbrace 1, 2, 4 \rbrace $ &  \\ [0.05cm]
Ionosphere & $\lbrace 1, 2 \rbrace $ &  \\ [0.05cm]
Iris & $\lbrace 1, 2, 3 \rbrace $&   \\  [0.05cm]
Satimage & $\lbrace 1, 3, 6, 9 \rbrace $~\cite{Gornitz2017} & \\ [0.05cm]
Segment & $\lbrace 1, 5, 7, 10, 14 \rbrace $~\cite{Gornitz2017}& \\
\bottomrule
\end{tabular}
\caption{Selected values for parameter $k$ (ClusterSVDD)}\label{Tuning}
\end{table}

\section{Results}\label{Results}
In this section, we first propose to the reader a preliminary experiment on synthetic data, to better understand the scope of the advocated method.  We then report the results achieved on benchmark datasets.
\subsection{Preliminary experiment on synthetic data}
We propose a first qualitative evaluation of our OC-Tree with ClusterSVDD with respect to the detection of three Gaussian blobs enclosing altogether 1000 instances. The parameter settings are given below. 
\begin{itemize}
\item OC-Tree : $\gamma = 0.05$, $\alpha = 1$, $\beta = 0\%$, $\nu = 0.1$.
\item ClusterSVDD : $k = 3$, $\nu_{SVDD} = 0.1$.
\end{itemize}   
The parameters of OC-Tree were established in a quite penalizing way, in the sense that setting $\alpha$ at $1$ means a systematic revision of any division with the risk of decomposing unnecessarily the space covered by the target instances. Moreover, setting $\beta$ at $0\%$ means no node is dropped; this may potentially lead to small hyper-rectangles to describe the target data.

\begin{figure}
\centering
\includegraphics[scale=0.55]{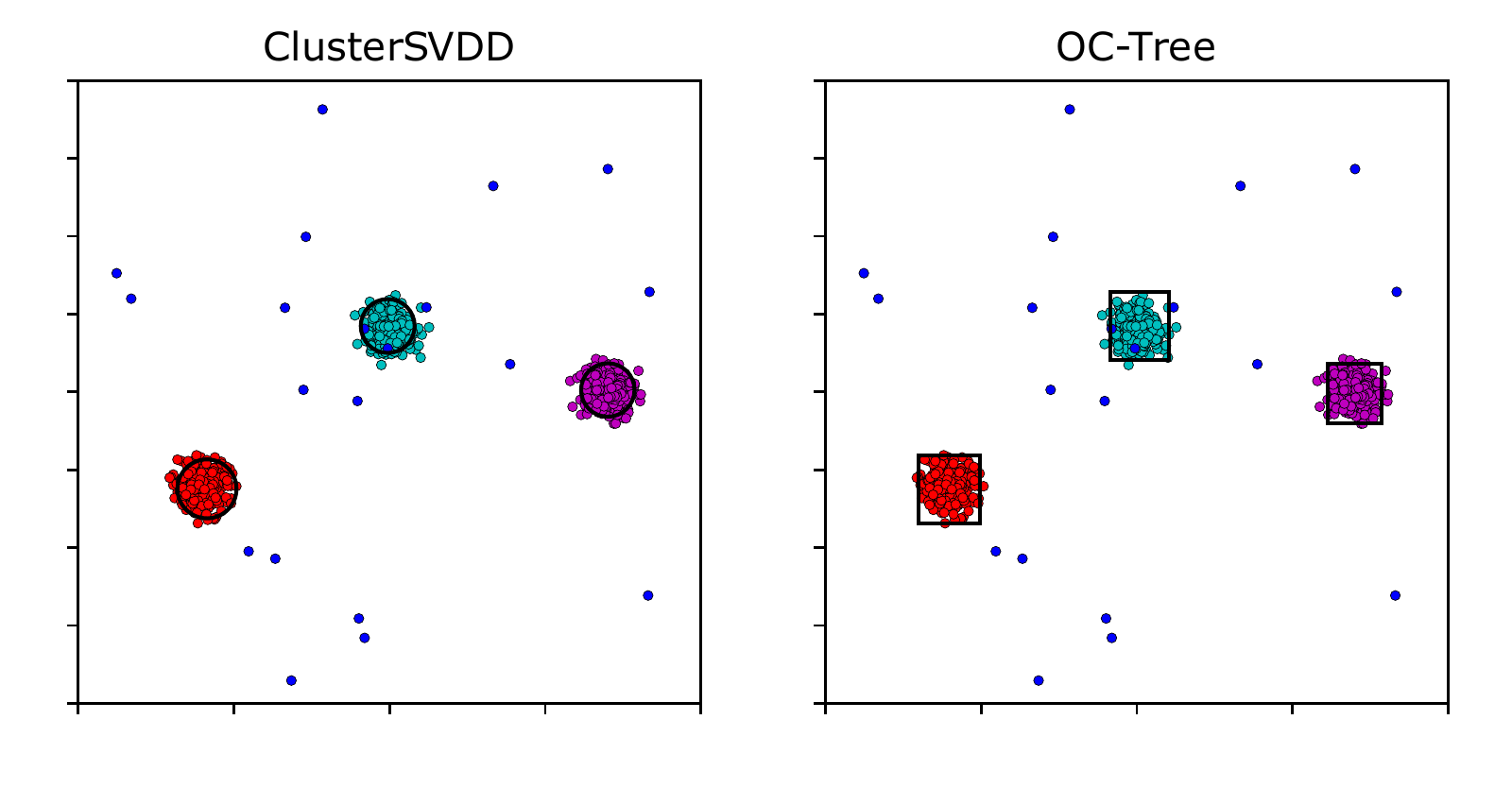}
\caption{Detection of three Gaussian blobs with 2\% of outliers included in the training set}\label{Exp1}
\end{figure}

\begin{figure}
\centering
\includegraphics[scale=0.55]{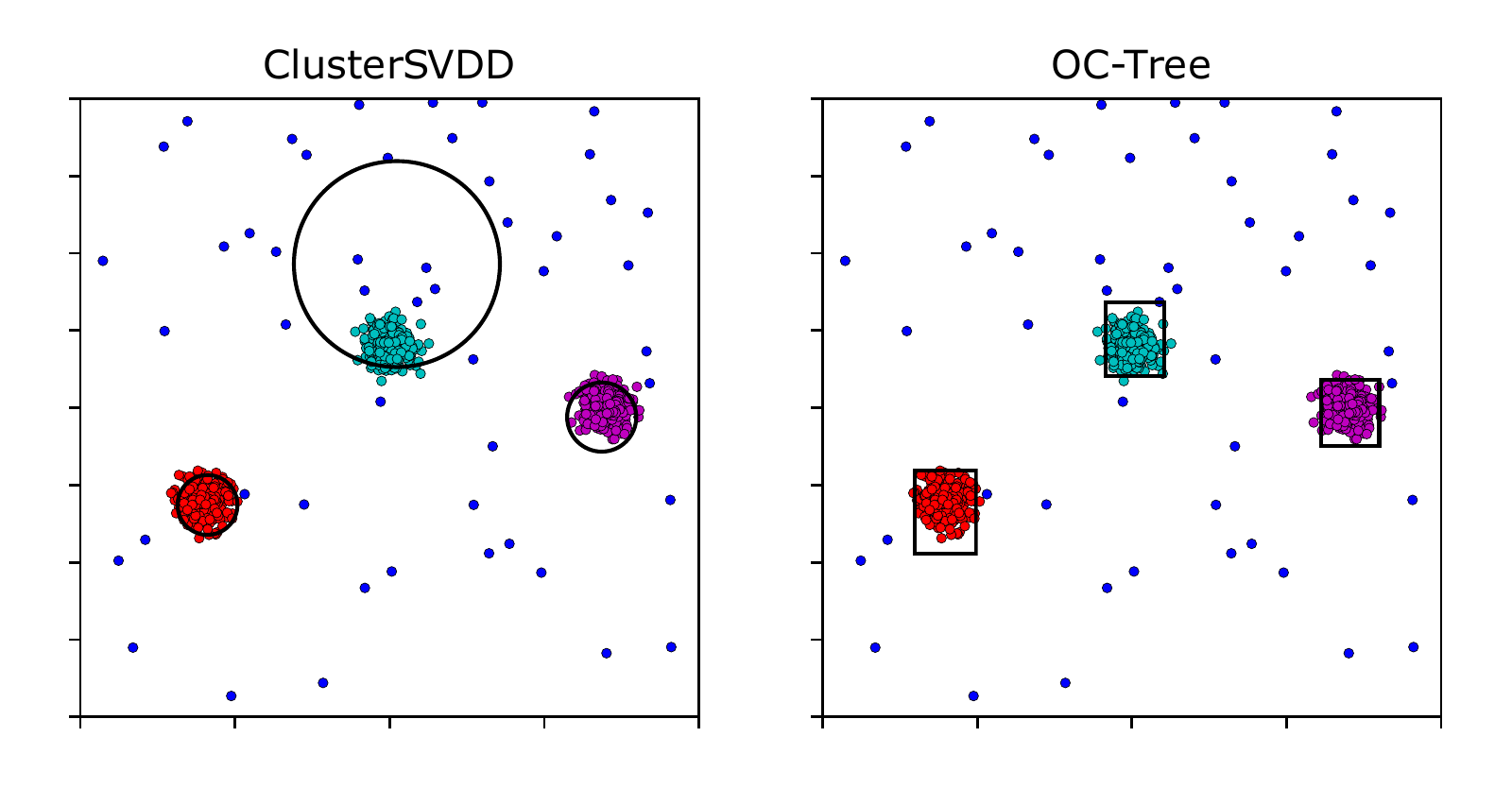}
\caption{Detection of three Gaussian blobs with 5\% of outliers included in the training set}\label{Exp2}
\end{figure}

Additional instances were added to the dataset in the form of uniformly distributed outliers, in proportions of $2\%$ and $5\%$ of the initial training set size respectively. The results are proposed in Figs.~\ref{Exp1} and~\ref{Exp2}. Both methods detect the blobs in the form of circles and rectangles respectively. However, it seems that OC-Tree is less sensitive to higher noise levels than ClusterSVDD (see Fig.~\ref{Exp2}).  Table~\ref{Perf1} compares the performances of ClusterSVDD and OC-Tree in terms of precision and recall. 

\begin{table}
\centering
\begin{tabular}{c|c|c|c|c}
\multirow{2}{*}{\textsc{Noise level}} & \multicolumn{2}{c|}{\textsc{ClusterSVDD}} & \multicolumn{2}{c}{\textsc{OC-Tree}}\\ \cmidrule{2-5}
&Precision& Recall & Precision & Recall \\ \toprule
2$\%$& 0.998 & 0.917 & 0.998 & 0.985 \\
5$\%$&0.995 & 0.940&0.999&0.987 \\
\bottomrule
\end{tabular}
\caption{Performance assessment on artificial data}\label{Perf1}
\end{table}

%
\subsection{Experiments on benchmark datasets}
In the present section, we compare our algorithm to ClusterSVDD, OCSVM, and iForest on benchmark datasets, according to the protocol summarized in Sec.~\ref{ExpPro}. Table~\ref{ExpA1} on the one hand, and Tables~\ref{ExpA2},~\ref{ExpA3} on the other hand summarize the results for the approaches A and B respectively. We report the couple of values Precision (P) -- Recall (R) for a validation achieved on the test sets. The results that are marked with an asterisk indicate that the corresponding method outperforms OC-Tree of more than 2\% in terms of F1-score. The lines that are succeeded with '(+)' indicates that the OC-Tree achieves a score superior or equal to the other techniques for the considered dataset.  
\subsubsection{Based on uniformly distributed noise -- Approach A}
Our first experiment was achieved on the benchmark datasets summarized in Table~\ref{Data}, in which uniformly distributed noise was injected in proportions of 2, 5, 10, 15~$\%$ of the initial dataset sizes. It appears that the OC-Tree performs favorably in comparison to the other reference methods. The improvements achieved against iForest may be explained by the fact that the latter method is properly intended for anomaly detection, and may thus have slightly lower performances when the proportion of outliers in the training set is low~\cite{Liu2008}, especially for proportions of 2\% and 5\%. Moreover, compared to iForest, the OC-Tree seems globally to better handle the \textit{ionosphere} and \textit{satimage} datasets. Actually, the \textit{ionosphere} dataset has a quite diffuse distribution of data along some dimensions, which involves that some normal instances may lie far away from the others. As it is built on a random choice of attributes, the iForest method is likely to detect these instances as outliers. On the opposite, the OC-Tree is built on attributes which concentrate the instances, so the ones lying outside these concentrations may be really perceived as outliers. As regards the \textit{satimage} dataset, the low proportion of outliers in such a high dimensional dataset may have disadvantaged the iForest method, with a difference in terms of F1-score that can reach 5\%. As regards the performances of OCSVM, they are in some cases lower than OC-Tree, which may be explained by the fact that OCSVM encloses data within a single boundary and can thus not exactly adjust to the structure of data. Finally, as mentioned previously, ClusterSVDD may be sensitive to noise, which explains why the OC-Tree provides better results in some cases.  

\begin{table*}
\centering
\resizebox{\textwidth}{!}{
\begin{tabular}{l|c|c|c|c||cc}
\textsc{Dataset} &  Noise level &ClusterSVDD & OCSVM & iForest & OC-Tree \\ \toprule 
\multirow{4}{*}{Australian} & 2\% &0.986 - 0.921& 0.995 - 0.908 & 1.000 - 0.926 & 1.000 - 0.965 & (+) \\
& 5\%& 0.973 - 0.926 & 0.977 - 0.922 & 1.000 - 0.922 & 1.000 - 0.970  & (+)\\ 
& 10\%& 0.900 - 0.960 & 0.916 - 0.960 & $\:\:$1.000 - 0.991*&0.900 - 0.996& \\ 
& 15\%& 0.877 - 0.961 & 0.882 - 0.935& 0.955 - 1.000&1.000 - 0.961& (+) \\ \midrule 
\multirow{4}{*}{Diabetes} & 2\%&1.000 - 0.941& 1.000 - 0.941 & 1.000 - 0.874  & 0.992 - 0.965 & (+) \\ 
&5\% & $\:\:$0.998 - 0.965* & 0.988 - 0.957 & 0.996 - 0.914 &0.992 - 0.911& \\ 
&10\%& 0.932 - 0.972 & 0.975 - 0.933 & 0.996 - 0.952 &0.980 - 0.952&\\ 
&15\% & 0.974 - 0.897 & 0.974 - 0.901&0.965 - 0.988 &0.992 - 0.945&\\ \midrule 
\multirow{4}{*}{Ionosphere} &2\% &0.972 - 0.914& 0.972 - 0.897  & 0.981 - 0.879 &1.000 - 1.000& (+)\\ 
&5\% & 0.938 - 0.913 & 0.937 - 0.904& 0.937 - 0.904 &1.000 - 1.000& (+) \\
&10\% & 0.884 - 0.939 & 0.880 - 0.904 & 0.904 - 0.912 & 0.983 - 1.000& (+) \\ 
&15\% & 0.828 - 0.946 & 0.824 - 0.920 & 0.832 - 0.929&0.982 - 1.000 & (+)\\ \midrule 
\multirow{4}{*}{Iris} &2\% &1.000 - 0.902& 1.000 - 0.961 & 1.000 - 0.922 & 1.000 - 0.941 &\\ 
&5\% & 0.977 - 0.860 & 0.980 - 0.960& 0.978 - 0.900&0.943 - 1.000 &(+)\\ 
&10\% & 0.979 - 0.920 & $\:\:$1.000 - 0.940*& 0.958 - 0.920& 0.978 - 0.900& \\ 
&15\% & 0.902 - 0.920 & 0.889 - 0.960& 0.889 - 0.960&0.862 - 1.000& (+)\\ \midrule 
\multirow{4}{*}{Satimage} &2\% &0.995 - 0.945& 0.996 - 0.957 & 0.996 - 0.890 & 0.996 - 0.968 &(+) \\ 
&5\% & 0.986 - 0.974 & 0.986 - 0.971 & 0.984 - 0.914& 0.979 - 0.981& (+)\\ 
&10\% & 0.991 - 0.981 & 0.977 - 0.946 & 0.952 - 0.922&0.981 - 0.969&\\ 
&15\% & 0.990 - 0.963 & 0.966 - 0.937& 0.907 - 0.934 &0.980 - 0.968&\\ \midrule 
\multirow{4}{*}{Segment} &2\% &0.999 - 0.963& 0.999 - 0.974  & 1.000 - 0.912  &1.000 - 1.000 &(+)\\ 
&5\% & 0.974 - 0.970 & 0.978 - 0.970& 1.000 - 0.940 & 1.000 - 1.000&(+)\\ 
&10\% & 0.927 - 0.980 & 0.930 - 0.979& 1.000 - 0.991 & 0.993 - 1.000&(+)\\ 
&15\% & 0.898 - 0.970 & 0.928 - 0.946 & $\:\:$0.976 - 1.000*&0.872 - 0.996& \\ 
\bottomrule
\end{tabular}}
\caption{Results - Approach A (P-R)}\label{ExpA1}  
\end{table*}

\begin{table*}
\centering
\resizebox{\textwidth}{!}{
\begin{tabular}{l|c|c|c|c||cc}
\textsc{Dataset} &  Noise level &ClusterSVDD & OCSVM & iForest & OC-Tree& \\ \toprule 
\multirow{4}{*}{Australian (-1)} & 2\% & 0.984 - 0.945 & 0.983 - 0.914& 0.991 - 0.875 & 0.992 - 0.977 & (+)\\ 
& 5\%& 0.936 - 0.936 &	0.935 - 0.920 &0.959 - 0.928& 0.945 - 0.960& (+)\\
& 10\%& 0.902 - 0.960 &0.919 - 0.912&0.941 - 0.896& 0.890 - 0.968& (+)\\ 
& 15\%&0.819 - 0.934&0.822 - 0.917&0.886 - 0.901&0.834 - 1.000& (+)\\ \midrule 
\multirow{4}{*}{Australian (+1)} & 2\% &0.980 - 0.951&0.980 - 0.951& 0.980 - 0.951 &0.990 - 0.980& (+)\\ 
& 5\%&0.950 - 0.950&0.950 - 0.941&0.949 - 0.921& 0.943 - 0.990& (+)\\ 
& 10\%&0.906 - 0.950&0.932 - 0.950&0.947 - 0.891&0.901 - 0.990& (+)\\ 
& 15\%&0.881 - 0.960&0.872 - 0.950&0.855 - 0.940& 0.860 - 0.980&\\ \midrule 
\multirow{4}{*}{Diabetes (-1)} & 2\%&  $\:\:$0.978 - 0.978* &$\:\:$0.977 - 0.966*&0.976 - 0.910& 0.976 - 0.910& \\
&5\% & $\:\:$0.957 - 0.978*&$\:\:$0.956 - 0.967*&$\:\:$0.954 - 0.922*&  0.952 - 0.878&\\ 
&10\%&0.944 - 0.934&0.926 - 0.956&0.928 - 0.846& 0.926 - 0.956& (+)\\ 
&15\% &0.863 - 0.921&0.876 - 0.955&0.874 - 0.933&0.862 - 0.910&\\ \midrule 
\multirow{4}{*}{Diabetes (+1)} & 2\%&  0.981 - 0.933&0.980 - 0.903&0.980 - 0.879&0.982 - 0.988& (+) \\ 
&5\% & 0.945 - 0.951 &0.956 - 0.927&0.955 - 0.909& 0.942 - 0.982& (+) \\ 
&10\%&0.895 - 0.962&0.905 - 0.956&0.909 - 0.938& 0.881 - 0.975& \\ 
&15\% &0.853 - 0.938&0.858 - 0.938&0.871 - 0.919&0.856 - 0.963& (+)\\ \midrule 
\multirow{4}{*}{Ionosphere (-1)} &2\% & 0.974 - 0.881 &0.967 - 0.690&0.971 - 0.810 & 0.977 - 1.000& (+)\\ 
&5\% & 0.946 - 0.833&0.935 - 0.690&0.946 - 0.833& 0.955 - 1.000& (+) \\ 
&10\% &0.872 - 0.829&0.857 - 0.732&0.872 - 0.829& 0.943 - 0.805& (+) \\ 
&15\% &0.889 - 0.930&0.861 - 0.721&0.895 - 0.791&0.905 - 0.884& (+)\\ \midrule 
\multirow{4}{*}{Ionosphere (+1)} &2\% &  1.000 - 0.960 &1.000 - 0.960&1.000 - 0.893& 0.986 - 0.973& (+)\\ 
&5\% &  0.973 - 0.973&0.986 - 0.946&0.986 - 0.919& 0.947 - 0.959&\\ 
&10\% &0.972 - 0.932&0.973 - 0.959&0.956 - 0.878&0.973 - 0.973& (+)\\ 
&15\% &$\:\:$0.920 - 0.958*&0.909 - 0.972&$\:\:$0.920 - 0.958*&0.861 - 0.944&\\ \midrule 
\multirow{4}{*}{Iris (1)} &2\% &$\:\:$1.000 - 1.000*&1.000 - 1.000&$\:\:$1.000 - 0.941*& 1.000 - 0.882& \\           
&5\% & 0.933 - 0.824&1.000 - 0.824&1.000 - 0.882& 1.000 - 0.882& (+)\\ 
&10\% &0.938 - 0.833&0.938 - 0.833&$\:\:$1.000 - 1.000*&1.000 - 0.889& (+)\\
&15\% &0.941 - 0.842&0.941 - 0.842&$\:\:$1.000 - 1.000*&1.000 - 0.895&\\ \midrule 
\multirow{4}{*}{Iris (2)} &2\% & 1.000 - 1.000 &1.000 - 1.000&1.000 - 0.824& 1.000 - 1.000&(+) \\       
&5\% & 0.944 - 1.000&0.944 - 1.000&0.944 - 1.000& 0.944 - 1.000&(+) \\
&10\% &1.000 - 1.000&0.947 - 1.000&1.000 - 1.000& 1.000 - 1.000&(+) \\
&15\% &$\:\:$1.000 - 1.000*&0.947 - 0.947&$\:\:$1.000 - 1.000*& 0.947 - 0.947& \\ \midrule  
\multirow{4}{*}{Iris (3)} &2\% & 1.000 - 0.824 &1.000 - 0.824&1.000 - 0.824& 1.000 - 1.000& (+) \\
&5\% & $\:\:$0.941 - 0.941*&0.933 - 0.824&0.933 - 0.824& 0.938 - 0.882& \\
&10\% &$\:\:$1.000 - 1.000*&1.000 - 0.722&$\:\:$1.000 - 1.000*& 1.000 - 0.833& \\
&15\% &0.929 - 0.684&1.000 - 0.789&1.000 - 0.895&1.000 - 0.895&(+) \\
\bottomrule
\end{tabular}}
\caption{Results (P-R) - Approach B}\label{ExpA2}  
\end{table*}

\begin{table*}
\centering
\resizebox{0.95\textwidth}{!}{
\begin{tabular}{l|c|c|c|c||cc}
\textsc{Dataset} &  Noise level &ClusterSVDD & OCSVM & iForest & OC-Tree \\ \toprule 
\multirow{4}{*}{Satimage (1)} &2\% & 0.997 - 0.975 &0.991 - 0.955&1.000 - 0.905& 0.997 - 0.958& \\
&5\% & 0.977 - 0.964  &0.980 - 0.972&0.997 - 0.952& 0.983 - 0.972&(+) \\
&10\% &0.958 - 0.992&0.955 - 0.986&0.986 - 0.981&0.970 - 0.992&(+) \\
&15\% &0.914 - 0.981&0.914 - 0.978&0.959 - 0.983&0.918 - 0.992& \\ \midrule
\multirow{4}{*}{Satimage (2)} &2\% & 0.980 - 0.942 &0.980 - 0.949 &0.986 - 0.872 & 0.979 - 0.910 &\\
&5\% & 0.994 - 0.981     &1.000 - 0.975&1.000 - 0.898& 0.980 - 0.955& \\
&10\% &0.910 - 0.987&0.915 - 0.981&0.967 - 0.942&0.937 - 0.968& \\
&15\% &0.925 - 0.948&0.902 - 0.955&$\:\:$0.950 - 0.987*&0.884 - 0.981&\\ \midrule
\multirow{4}{*}{Satimage (3)} &2\% &  0.984 - 0.984 &0.984 - 0.981&1.000 - 0.953& 0.987 - 0.959&\\
&5\% & 0.984 - 0.972 &0.984 - 0.966&0.994 - 0.957&  0.966 - 0.975&\\
&10\% &0.942 - 0.985&0.944 - 0.988&0.979 - 0.985&0.972 - 0.948& \\
&15\% &0.906 - 0.981 &0.917 - 0.985&0.944 - 0.988&0.924 - 0.971& \\ \midrule
\multirow{4}{*}{Satimage (4)} &2\% &  0.984 - 0.933 &0.984 - 0.933&0.992 - 0.926& 0.984 - 0.933 & (+)\\
&5\% & 0.969 - 0.941  &0.977 - 0.941&0.992 - 0.889& 0.964 - 0.985 & (+)\\
&10\% &0.907 - 0.948&0.920 - 0.948&0.961 - 0.918&0.917 - 0.985& (+) \\
&15\% &0.869 - 0.940&0.910 - 0.978&0.928 - 0.955&0.905 - 0.993&(+) \\ \midrule
\multirow{4}{*}{Satimage (5)} &2\% &  0.980 - 0.948 &0.979 - 0.935&0.993 - 0.869& 0.966 - 0.941& \\
&5\% & 0.967 - 0.961  &0.966 - 0.947&0.993 - 0.882& 0.931 - 0.974& \\
&10\% &0.937 - 0.955&0.932 - 0.968&0.942 - 0.929&0.921 - 0.968& \\
&15\% &0.876 - 0.993&0.883 - 0.953&0.898 - 0.940&0.865 - 9.980& \\ \midrule 
\multirow{4}{*}{Satimage (6)} &2\% &  0.997 - 0.977 &0.994 - 0.977&1.000 - 0.951& 0.994 - 0.986&(+) \\  
&5\% & 0.977 - 0.968 &0.977 - 0.971&0.988 - 0.942& 0.980 - 0.991&(+) \\
&10\% &0.948 - 0.986&0.950 - 0.986&0.977 - 0.980&0.961 - 0.980& \\
&15\% &0.910 - 0.972&0.931 - 0.952&0.941 - 0.997& 0.966 - 0.957& \\ \midrule
\multirow{4}{*}{Segment (1)} &2\% &   $\:\:$0.982 - 0.973* &$\:\:$0.982 - 0.982*&1.000 - 0.855 & 1.000 - 0.873& \\
&5\% & 0.981 - 0.972   &0.972 - 0.981&1.000 - 0.906& 1.000 - 0.934& \\
&10\% &0.915 - 1.000&0.938 - 0.991&$\:\:$1.000 - 1.000*&0.945 - 0.972& \\
&15\% &0.945 - 0.963&0.938 - 0.981&$\:\:$0.973 - 1.000*& 0.919 - 0.953& \\ \midrule
\multirow{4}{*}{Segment (2)} &2\% &  0.991 - 0.955 &0.990 - 0.936&1.000 - 0.882 & 1.000 - 0.927& \\
&5\% &  0.981 - 0.972   &0.981 - 0.962&1.000 - 0.934& 0.955 - 0.991&(+) \\
&10\% &0.910 - 0.944&$\:\:$0.927 - 0.944*&1.000 - 0.972&0.955 - 0.991& \\
&15\% &0.855 - 0.991&$\:\:$0.851 - 0.963*&$\:\:$0.964 - 1.000*&0.990 - 0.897& \\ \midrule
\multirow{4}{*}{Segment (3)} &2\% &   0.981 - 0.918 &0.981 - 0.936&0.990 - 0.927& 1.000 - 0.982&(+) \\
&5\% & 0.950 - 0.896 &0.949 - 0.887&0.939 - 0.877& 0.962 - 0.943&(+)  \\
&10\% &0.909 - 0.935&0.925 - 0.925&0.951 - 0.916&0.904 - 0.972&(+) \\
&15\% &0.862 - 0.935&0.860 - 0.916&0.907 - 0.907&0.866 - 0.963 &(+)\\ \midrule
\multirow{4}{*}{Segment (4)} &2\% &  1.000 - 0.945 &1.000 - 0.955&0.990 - 0.918& 0.991 - 1.000&(+) \\
&5\% &  0.952 - 0.934  & 0.981 - 0.962&0.990 - 0.925&  0.981 - 0.991&(+) \\
&10\% &0.937 - 0.972&0.945 - 0.963&0.971 - 0.925&0.922 - 0.991&(+)\\
&15\% &0.898 - 0.991&0.906 - 0.991&0.927 - 0.953&0.869 - 0.991& \\ \midrule
\multirow{4}{*}{Segment (5)} &2\% &  0.981 - 0.945 &0.981 - 0.964&0.990 - 0.927 & 0.991 - 0.964&(+) \\
&5\% & 0.952 - 0.934&0.962 - 0.962&0.963 - 0.972& 0.955 - 0.991&(+) \\
&10\% &0.917 - 0.925&0.920 - 0.963&0.936 - 0.963&0.921 - 0.981                     &(+) \\
&15\% &0.864 - 0.953&0.858 - 0.963&$\:\:$0.938 - 0.981*&0.862 - 0.991& \\ \midrule
\multirow{4}{*}{Segment (6)} &2\% &  0.991 - 0.964 &0.991 - 0.964&1.000 - 0.909& 0.990 - 0.936& \\
&5\% & 0.971 - 0.943&0.980 - 0.934&0.990 - 0.925&                               0.981 - 0.953&(+) \\
&10\% &0.955 - 0.991&0.964 - 0.991&$\:\:$0.981 - 0.991*& 0.930 - 0.991& \\
&15\% &0.946 - 0.981 &0.946 - 0.981&$\:\:$0.964 - 0.991*& 0.898 - 0.991& \\ \midrule 
\multirow{4}{*}{Segment (7)} &2\% &  1.000 - 0.918&1.000 - 0.909&1.000 - 0.945& 1.000 - 0.982&(+) \\
&5\% &0.970 - 0.925&0.970 - 0.906&1.000 - 0.925&1.000 - 0.972&(+) \\
&10\% &0.909 - 0.935 &$\:\:$0.919 - 0.953*&$\:\:$1.000 - 0.981*& 0.886 - 0.944& \\
&15\% &0.863 - 0.944&0.871 - 0.944&$\:\:$0.939 - 1.000*& 0.875 - 0.981& \\
\bottomrule
\end{tabular}}
\caption{Results (P-R) - Approach B}\label{ExpA3}  
\end{table*}

\subsubsection{Based on the one vs rest strategy -- Approach B}
In this case, the multi-class problems related to the considered datasets are converted to one-class problems in which the representatives of the other classes are considered as outliers, injected in proportions of 2, 5, 10, 15~$\%$ of the one-class dataset sizes. In such a situation, we can expect that reference methods such as OCSVM and iForest perform better since they handle the data of each class separately. De facto, the OC-Tree shows overall smaller differences in performance. 
%


\section{Application to the diagnosis of ADHD}\label{CaseStudy}
In the previous section, we compared the OC-Tree on benchmark datasets with reference one-class methods, against which it proved to perform favorably. In the present section, we propose a real-world case study in which an algorithm such as the OC-Tree is worth considering.  The application is related to the diagnosis of Attention Deficit Hyperactivity Disorder (ADHD). 
\subsection{Problem statement}
ADHD is a neurodevelopmental disorder in children which has been subject to a considerable number of studies, including those conducted on the ADHD-200 collection~\cite{Bellec2017}. This open and free database has been made available since 2012 in order to advance the state of knowledge about ADHD~\cite{Milham2012adhd}. 

The epidemiology of ADHD depends on gender, and evidence suggests that the disorder affects more often boys than girls~\cite{ADHDinstitute}. Such a gender-differentiated distribution poses some concerns about the development of diagnosis aid models through multi-class classification. Indeed, unbalanced distributions of ADHD and NeuroTypical (NT) subjects are often observed for each gender group in the training sets related to ADHD. This applies to the ADHD-200 collection, and more particularly to the corresponding NYU data subset. The boys' training sample includes approximately twice as many ADHD subjects as NT ones, and the reverse trend is observed in the girls' training sample. Fig.~\ref{ConfMat} presents the confusion matrices related to the predictions recently achieved on the NYU test set for boys and girls, based on a multi-class decision tree (according to the methodology proposed in~\cite{Itani2019b}). Actually, these results show the effects of class unbalance within each gender group in the training set. Though providing an overall satisfactory predictive accuracy, the final predictive model has a high (resp. low) sensitivity and a low (resp. high) specificity in boys (resp. girls).

\begin{figure}
\centering
\resizebox{0.9\textwidth}{!}{
   \begin{minipage}[c]{.4\linewidth}
			\renewcommand{\arraystretch}{1.2}
			\centering
			      \begin{tabular}{|c|c|c|}
			         \hline
			            Predicted as $\triangleright$  &NT &  ADHD \\ \hline 
			            NT &\color{ForestGreen}3\color{black}& \color{red}5\color{black} \\ \hline
			           ADHD&\color{red}1\color{black}&\color{ForestGreen}19\color{black} \\ \hline 
			      \end{tabular}
   \end{minipage} \hspace{2.2cm}
   \begin{minipage}[c]{.4\linewidth}
\renewcommand{\arraystretch}{1.2}
\centering
      \begin{tabular}{|c|c|c|}
         \hline
           Predicted as $\triangleright$  &NT &   ADHD \\ \hline 
           NT &\color{ForestGreen}4\color{black}&\color{red}0\color{black}\\ \hline
           ADHD &\color{red}5\color{black}&\color{ForestGreen}4\color{black}\\ \hline 
      \end{tabular}
   \end{minipage}}
   \caption{Confusion matrices achieved on the NYU test set for boys (left) and girls (right), based on a multi-class decision tree~\cite{Itani2019b}}\label{ConfMat}
\end{figure}
\noindent
This bias is among the reasons that explain the limited applicability of such a binary predictive model in the clinical practice setting. The OC-Tree may alleviate this issue. We thus propose to tackle ADHD diagnosis on a gender-differentiated basis, in focusing on the description of the neuropathology with the OC-Tree. 

\subsection{Data}

We consider the preprocessed ADHD-200 collection~\cite{AthenaPipeline}, and focus on the NYU sample. Table~\ref{DistADHD} presents the distribution of the training and test data, based on the gender and the diagnostic labels. For each subject, the sample includes blood-oxygen-level-dependent signals~\cite{Aguirre1998}, at resting-state, given a brain parcellation in 90 regions of interest (cf. AAL90 atlas~\cite{Tzourio2002}). We considered the variance of the signals as predictors, since they proved to achieve successful predictions~\cite{Itani2018, Itani2019b}. They were computed for brain regions included in two functional systems which were associated to ADHD-related abnormalities in the literature: the limbic system~\cite{Itani2019b} and the Default Mode Network (DMN)~\cite{Tamm2012, Broyd2009}.

\begin{table}
\centering
\begin{tabular}{llcc}
&& Girls & Boys \\ \toprule
\multirow{2}{*}{Training set}&NT & 50   &43 \\
&ADHD& 25  & 92  \\ \midrule
\multirow{2}{*}{Test set}&NT  &4&8 \\ 
&ADHD&9&20 \\ \bottomrule
\end{tabular}
\caption{Distribution of the NYU sample considered in our study}\label{DistADHD}
\end{table}

\subsection{Tuning and assessment}

\begin{figure}
\centering
\includegraphics[scale=0.7]{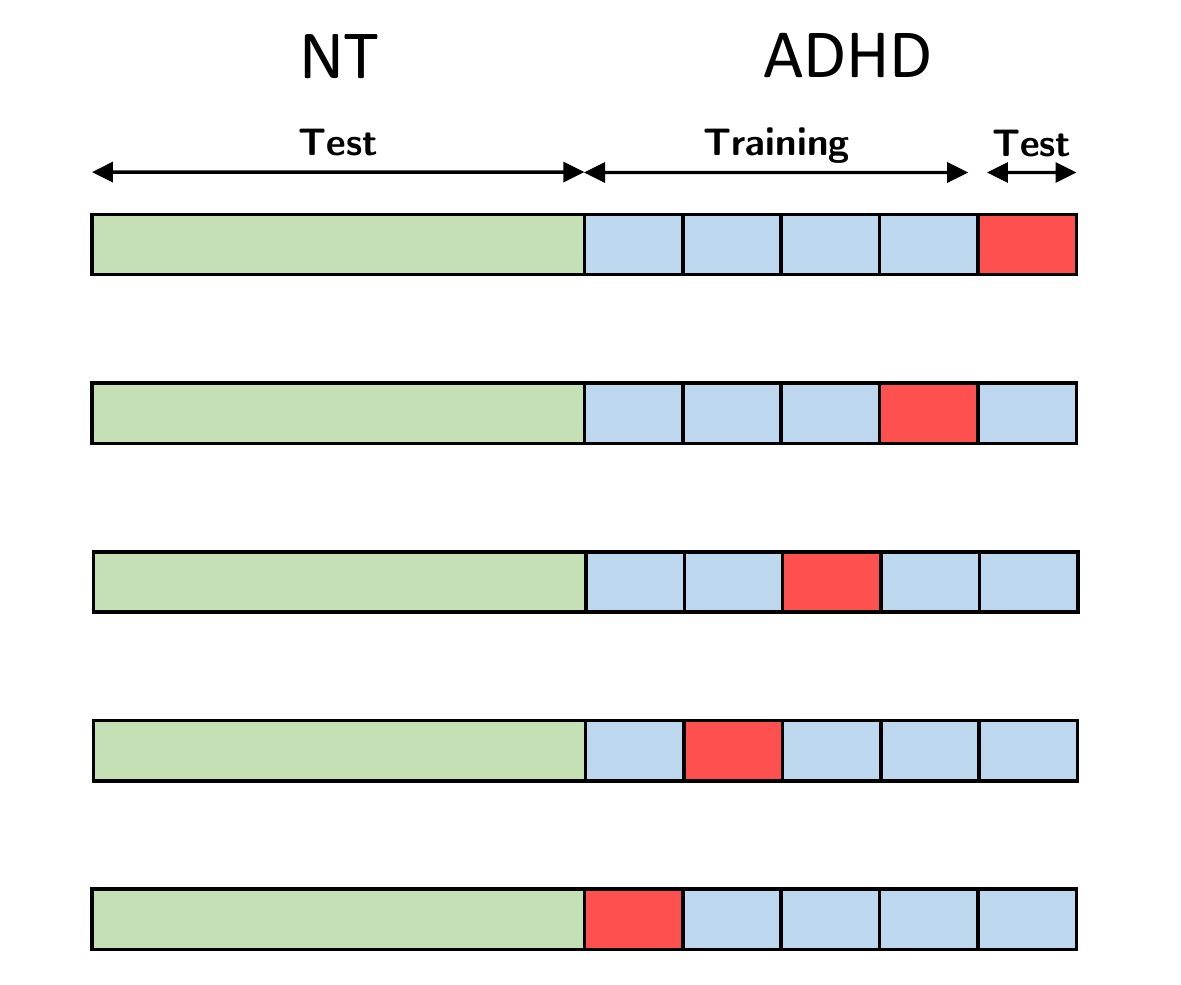}
\caption{Cross-validation procedure used to tune the OC-Tree for ADHD prediction}\label{CV2}
\end{figure}

In this context, a quick visualization of the data shows that the instances are concentrated within a single grouping. Thus, there are no clusters to raise: the models may be reduced to a set of descriptive rules. This means that the parameter $\alpha$ has no influence here.  Five values were considered in order to tune parameter $\nu = \lbrace 0.05, 0.1, 0.15, 0.2, 0.25 \rbrace$. The parameter was tuned through a 5-fold CV procedure, which is depicted in Fig.~\ref{CV2}. The NT subjects of the training set are fully used at each iteration as a test fold in combination to the one extracted from the partitioning of the ADHD training set into 5 folds. 


In OCC, performance metrics such as those presented in  Sec.~\ref{AssessmentMet} are generally computed with regards to the target class, i.e. ADHD in this case.  However, in the specific case of psychiatric diagnosis, there is a need for cautious predictions, even though that would imply to wrongly predict a subject as neurotypical~\cite{Itani2019a}. In other terms, high specificity and a reasonable level of sensitivity are requirements that a predictive model should meet in this context. We thus propose to assess the model towards its capability to predict NT cases, and thus to compute the metrics with respect to the NT group.

The models which achieve the best F1-score and precision were held as relevant for boys and girls respectively. Indeed, let us recall that our choice to assess the performance of the OC models towards the class of typical controls is motivated by the need to favor high levels of specificity. However, this is an already existing trend in girls, given that there are generally more NT girls than ADHD ones. Thus, to avoid falling into the traps of a somewhat insensitive model and to ensure that ADHD cases are predicted in a reasonable number of situations, we focus on the precision whose maximization is achieved through a minimization of the number of false NT subjects.

\subsection{Classification framework}

On a gender-differentiated basis, we need to predict a diagnosis based on the activity of brain regions included in the limbic system and/or the DMN. As announced, this is achieved through the practice of OCC, in targeting the ADHD group. For such a purpose, we assessed the relevance of four distinct options presented below.  
\begin{itemize}
\item \textbf{O1}: train the OC-Tree model on the features related to the limbic system.
\item \textbf{O2}: train the OC-tree model on the features related to the DMN. 
\item \textbf{O3}: train the OC-tree model on features related to both the limbic system and DMN.
\item \textbf{O4}: constitute an ensemble of OC classifiers by the aggregation of two models trained on the limbic and DMN features separately. 
\end{itemize}
In the case of the fourth option, a subject is diagnosed with ADHD once he/she tests positive with both models. In the other cases, the subject is predicted as disease-free, concerned with the need for a cautious diagnosis~\cite{Itani2019a}. 

\subsection{Final models and performance}

In boys, the ensemble strategy as defined by option O4 appeared to be the most successful, with a F1-score of 65.3\% on the training set ($\nu$ = 0.25). The results were most tightly contested in girls between option O1 and O3, yielding respectively precision rates of 93.6\% and 93.3\% ($\nu$ = 0.15). We selected the latter as a final model since it provides a more detailed description of the pathology than O1, which is based only on two rules. Fig.~\ref{CMfinal} presents the final confusion matrices for boys and girls.

Tables~\ref{ModelBoys} and~\ref{ModelGirls} present the decision rules (expressed in terms of the logarithm of the variance) related to boys and girls respectively. Note that (L) and (R) denote brain regions included in the Left and Right hemispheres respectively. Our results confirm that the resting-state activity of both the limbic system and the DMN brings some discriminative information for ADHD diagnosis. The mental condition appears to be more complex to describe in boys, and requires the combination of two distinct models. The girls' model is by contrast more minimalist. These important differences between boys and girls in terms of models strengthens our conviction that a gender-differentiated classification is definitely pertinent. 

\begin{figure}
\centering
\resizebox{0.9\textwidth}{!}{
   \begin{minipage}[c]{.4\linewidth}
			\renewcommand{\arraystretch}{1.2}
			\centering
			      \begin{tabular}{|c|c|c|}
			         \hline
			           Predicted as $\triangleright$ &NT & ADHD\\ \hline 
			            NT &\color{ForestGreen}6\color{black}& \color{red}2\color{black}\\ \hline
			           ADHD&\color{red}6\color{black}&\color{ForestGreen}14\color{black}\\ \hline 
			      \end{tabular}
   \end{minipage} \hspace{2.2cm}
   \begin{minipage}[c]{.4\linewidth}
\renewcommand{\arraystretch}{1.2}
\centering
      \begin{tabular}{|c|c|c|}
         \hline
           Predicted as $\triangleright$ & NT&ADHD \\ \hline 
            NT&\color{ForestGreen}4\color{black}&\color{red}0\color{black}\\ \hline
          ADHD &\color{red}1\color{black}&\color{ForestGreen}8\color{black}\\ \hline 
      \end{tabular}
   \end{minipage}}
         \caption{Confusion matrix achieved by the OC-Tree on the NYU test set for boys (left) and girls (right)}\label{CMfinal}
\end{figure}

In alleviating the issue of class imbalance within each gender group, we could improve the balance between the diagnostic specificity and sensitivity. If we compare with the confusion matrices presented in Figs.~\ref{ConfMat} and \ref{CMfinal}, in boys, the improvement made on specificity (75\% against 37.5\% previously), was achieved at the expense of the sensitivity (70\% against 95\% previously). In girls, the sensitivity was doubled without loss of specificity. The overall prediction accuracy was improved as well (78.0\% against 73.2\%).

 \begin{sidewaystable}
 \renewcommand{\arraystretch}{1.2}
 \small
 \resizebox{\textwidth}{!}{
 \begin{tabular}{lp{4cm}|lp{4cm}}
\multicolumn{2}{c|}{\textsc{Model 1 - Limbic system}} & \multicolumn{2}{c}{\textsc{Model 2 - DMN}} \\
\textbf{Brain region} &  \multicolumn{1}{p{1cm}|}{\textbf{log(Var)}} & \textbf{Brain region} & \textbf{log(Var)} \\
 Parahippocampal (L) & [-2.28 ; -1.33]& Angular gyrus (R) & [-2.09 ; -1.29]\\
Posterior cingulate gyrus (L) & [-1.49 ; -0.81] & Parahippocampal (L)& [-2.24 ; -1.42]\\
Amygdala (R) & [-2.17 ; -1.48]& Superior frontal gyrus, dorsolateral (L)& [-2.43 ; -1.59]\\
 Superior frontal gyrus, dorsolateral (L) &[-2.43 ; -1.60]&Superior frontal gyrus, medial& [-2.08 ; -1.23] \\
Hyppocampus (R) &[-2.61 ; -1.81]&Posterior cingulate gyrus (L) & [-1.49 ; -0.81]\\
Anterior cingulate and paracingulate gyri (R) &[-2.54 ; -1.55]& Angular gyrus (L)& [-1.83 ; -1.12]\\
Anterior cingulate and paracingulate gyri (L) &[-1.92 ; -1.04]&Superior frontal gyrus, medial (L)& [-1.88 ; -0.97]\\
Superior frontal gyrus, dorsolateral (R) &[-2.61 ; -1.70]&& \\
Thalamus (R) &[-2.46 ; -1.57]&& \\
Posterior cingulate gyrus (R) &[-1.84 ; -1.03]&& \\ \bottomrule 
\end{tabular}   }
\caption{Predictive models for boys}\label{ModelBoys}
\vspace{1cm}
\centering
 \resizebox{0.4\textwidth}{!}{
\begin{tabular}{lp{2.5cm}}
\textbf{Brain region} & \textbf{log(Var)} \\ \toprule 
Middle temporal gyrus (R) & [-2.28 ; -1.85] \\
Thalamus (L) & [-2.43 ; -1.67]\\
Thalamus (R) & [-2.18 ; -1.40] \\ \bottomrule
\end{tabular}}
\caption{Predictive model for girls}\label{ModelGirls}
 \end{sidewaystable}

\section{Discussion}\label{Discussion}

In Sec.~\ref{Results}, we showed that the OC-Tree presents favorable performances in comparison to reference methods such as ClusterSVDD, OCSVM, and iForest, in similar conditions. Depending on the targeted objectives, the OC-Tree may be a wise choice to achieve a OCC task. 

The presence of noise in the data may impair the performances of a method like ClusterSVDD, while as a density-based method, the OC-Tree shows more ability to reject such outliers in the data. Moreover, the OC-Tree is developed to be as compact as possible, which constitutes a key to interpretability. Indeed, the predictive model is based on the most discriminative attributes to achieve OCC while ClusterSVDD and OCSVM do not consider such a selection; the corresponding models are computed based on the whole set of training attributes. The OC-Tree also detects automatically the number of groupings related to the class targeted by the classification. This constitutes a significant advantage compared to ClusterSVDD which requires to set the number of possible clusters as an input parameter. As compared to the iForest technique, the OC-Tree is more compact and readable while being able at the same time to perform outlier rejection. Finally, the OC-Tree better fits to the structure of the data as compared to OCSVM, since it allows the detection of sub-concepts of a single class as target groupings.

In Sec.~\ref{CaseStudy}, we were interested in a case study related to the diagnosis of ADHD. Through this study, we could: 
\begin{itemize}
\item show the interest of considering the OC-Tree rather than a multi-class decision tree, given the effective availability of the data ; 
\item highlight the advantageous interpretability of the OC-Tree, which is an important characteristic towards a concrete clinical applicability; 
\item consider one-class ensembles that may help in modeling complex conditions while preserving the interpretability of the  predictive framework. 
\end{itemize}
These promising results tend to show that our model may be transposable to medical practice as a diagnosis aid tool.

\section{Conclusion \& future work}\label{Conclusion}
In some applications, the limited availability of data has lead to look for alternatives to the traditional supervised techniques. The practice of One-Class Classification (OCC) has been considered in this context. This area of machine learning has generated a considerable interest with the development of new methods, some of which were adapted from supervised classification techniques.

In this work, we proposed a one-class decision tree by completely rethinking the splitting mechanism considered to build such models. Our One-Class Tree (OC-Tree) may be actually seen as an adaptation of the Kernel Density Estimation (KDE) for the sake of interpretability, based on a subset of significant attributes for the purpose of prediction. In that respect, our method has shown favorable performances in comparison to reference methods such as ClusterSVDD, one-class SVM and isolation forest. Against these approaches, our one-class model is quite simple while being in the same time transparent and performant. Such qualities are particularly valuable for medical diagnosis, where a balanced representation of the  classes is not always ensured. We could illustrate the benefits of the OC-Tree for the diagnosis of ADHD. Our results show that the OC-Tree constitutes a step towards greater applicability of diagnosis aid models. 

This work leaves some interesting perspectives. In particular, the parametrization of the KDE remains an open question as regards the computation of the bandwidth $h$ and the use of other kernels $K$. Indeed, on the one hand, our proposal is based on a Gaussian kernel attractive by its mathematical properties, but the pertinence of other configurations may be studied on a comparative basis. On the other hand, deduced based on the Silverman's rule of thumb, $h$ is quite sensitive to the training set content. In our proposal, this sensitivity is controlled by setting a pre-pruning mechanism. In the future, we would like to rise to the challenge of establishing a rule able to address this issue of sensitivity.

\section{Acknowledgments}
Sarah Itani is a research fellow of Fonds de la Recherche Scientifique - FNRS (F.R.S.- FNRS).

\end{document}